\documentclass[10pt,conference,a4paper]{IEEEtran}

\usepackage{graphics} % for pdf, bitmapped graphics files
\usepackage[]{algorithm2e}
\usepackage{epsfig} % for postscript graphics files
\usepackage{mathptmx} % assumes new font selection scheme installed
\usepackage{amsmath} % assumes amsmath package installed
\usepackage{balance}
\usepackage{multirow}

\usepackage{makecell}
\usepackage[table]{xcolor}

\ifCLASSINFOpdf
\else
\fi
\begin{document}
\title{Learning from Learners: Adapting Reinforcement Learning Agents to be Competitive in a Card Game.}

\author{\IEEEauthorblockN{Pablo Barros, Ana Tanevska and Alessandra Sciutti}
\IEEEauthorblockA{Cognitive Architecture for \\Collaborative Technologies(CONTACT) Unit \\Istituto Italiano di Tecnologia\\ Genova, Italy\\
Email: \{pablo.alvesdebarros, ana.tanevska, alessandra.sciutti\}@iit.it}
% \and
% \IEEEauthorblockN{Ana Tanevska}
% \IEEEauthorblockA{Cognitive Architecture for \\Collaborative Technologies\\ (CONTACT) Unit \\Istituto Italiano di Tecnologia\\ Genova, Italy
% Email: ana.tanevska@iit.it}
% \and
% \IEEEauthorblockN{Alessandra Sciutti}
% \IEEEauthorblockA{Cognitive Architecture for \\Collaborative Technologies\\ (CONTACT)\ Unit \\Istituto Italiano di Tecnologia\\ Genova, Italy
% Email: alessandra.sciutti@iit.it}

}


%Telephone: (800) 555--1212, Fax: (888) 555--1212}


\maketitle

\begin{abstract}
Learning how to adapt to complex and dynamic environments is one of the most important factors that contribute to our intelligence. Endowing artificial agents with this ability is not a simple task, particularly in competitive scenarios. In this paper, we present a broad study on how popular reinforcement learning algorithms can be adapted and implemented to learn and to play a real-world implementation of a competitive multiplayer card game. We propose specific training and validation routines for the learning agents, in order to evaluate how the agents learn to be competitive and explain how they adapt to each others' playing style. Finally, we pinpoint how the behavior of each agent derives from their learning style and create a baseline for future research on this scenario.
\end{abstract}


% \fi
\IEEEpeerreviewmaketitle

\section{Introduction}

% Learning how to adapt to complex and dynamic environments is one of the most important factors that contribute to our intelligence. Translating this characteristic into artificial agents proved to be a difficult task \cite{mnih2015human, hessel2018rainbow, henderson2018deep}. The mapping between actions and environmental rewards became a very popular solution for creating autonomous artificial agents that do not rely on explicit supervision, in particular with the last two decades of development on reinforcement learning. Such agents usually embed different policy learning mechanisms that allow them to develop decision-making strategies which are, in some cases, unexpected or unknown by humans \cite{wang2016does, silver2018general}. 

% With the recent interest in reinforcement learning caused by the development of deep reinforcement learning techniques \cite{mnih2013playing}, novel methods, mechanisms, and scenarios were developed in recent years. Such mechanisms allow the agent to process highly complex scenarios, such as high-resolution raw images \cite{lillicrap2015continuous}, in an end-to-end learning manner reducing the need for strong and well-defined prior knowledge.

% The proliferation of different reinforcement learning techniques contributes to the incredible increase on popularity that such research field had recently, which as a consequence p

With the current interest in reinforcement learning caused by the development of deep reinforcement learning techniques \cite{mnih2013playing}, novel methods and mechanisms have been developed in recent years. Such mechanisms allow an artificial agent to map between state and actions within highly complex state representations and in an end-to-end learning manner, reducing the need for strong and well-defined prior knowledge. In recent cases, reinforcement learning agents have been used for guiding autonomous cars \cite{sallab2017deep, isele2018navigating}, predicting the stock exchange impact \cite{ponomarev2019using, meng2019reinforcement}, and coordinating a swarm of robots to protect the environment \cite{haksar2018distributed, yu2018reinforcement}. %, such as high-resolution raw images \cite{lillicrap2015continuous},

Most of these solutions, although having real-world-inspired scenarios, focus on a direct space-action-reward mapping between the agent's action and the environment state. That translates to agents that can adapt to dynamic scenarios, but, when applied to competitive scenarios, they fail to address the impact of the opponents. In most cases, when these agents choose an action, they do not take into consideration how other agents can affect the state of the scenario. In this regard, competitive reinforcement learning is still behind the mainstream applications and demonstrations of the last years.

In competitive scenarios, the agents have to learn decisions that a) maximize their goal, and b) minimize their adversaries' goals. Besides dealing with complex scenarios, they usually have to deal with the dynamics between the agents themselves. Some of the most common applications for competitive reinforcement learning involve multi-agent simulations, such as multiple autonomous vehicles \cite{fridman2018deeptraffic}, life-simulation/resources gathering \cite{xu2018hierarchical}, pursuer/pursued scenarios \cite{wang2019competitive}), and multi-player games \cite{mckenzie2017competitive}. 

The recent development and popular interest in deep reinforcement learning have contributed, however, to the design, implementation, and evaluation of only a few competitive learning solutions. The implementation of a counterfactual thinking solution \cite{wang2019competitive}, based on a classic psychological phenomenon, obtained a good performance on a simple multi-agent resource gathering life-simulation water world scenario \cite{gupta2017cooperative}. The model is certainly interesting but became very complex to scale to realistic scenarios as it implements an extra counterfactual policy network that is extremely sensitive to hyperparameters change. In another direction, a centralized learning mechanism was introduced by Tampuu et al.  \cite{tampuu2017multiagent}. This presents an effective way of learning competitive actions, but it demands the learner to have total control of the environment, which restricts its applications. Moreover, all of these models were evaluated using very limited simulations of real-world events and most of the time do not scale well to real-world problems \cite{hernandez2019survey}.

To better assess how popular reinforcement learning methods perform in a real-world competitive scenario, we propose a broad study on how different reinforcement learning agents learn and behave when deployed in such an environment. We investigate how three reinforcement learning models (Deep Q-Learning - DQL \cite{van2016deep}, Advantage Actor-Critic - A2C \cite{mnih2016asynchronous}, and Proximal Policy Optimization - PPO \cite{schulman2017proximal}) can learn a competitive multiplayer card game, and evaluate how their emerged behavior affect their own decisions towards winning the game. By focusing on these three implementations, we aim to provide the training, analysis and performance baseline for the competitive Chef's Hat card game \cite{barros2020food}, without the need of a centralized learner or overly-complex solutions. Our goal is to understand how these established models behave in a real-world inspired competitive scenario.

To maintain our scenario as close to real-world as possible, we implement in full the Chef's Hat card game, which has been designed to be used in Human-Robot Interactions (HRI). The game contains specific mechanics that allow complex dynamics between the players to be used in the development of a winning game strategy. We use the OpenAI Gym-based Chef's Hat simulation environment \cite{barros2020chef} to emulate, in a 1:1 scale, all the possible game mechanics. A card game scenario allows us to have a naturally-constrained environment and yet obtain responses that are the same as the real-world counter-part application. It additionally helps us to better understand the decision-making process of the agents and to better illustrate the strategies learned by each agent and how they affect each other.

For each of the three reinforcement learning methods, we introduce adaptations to the learning mechanisms of each agent, including a novel greedy policy for action selection. We perform three main competitive learning tasks: first, each of these agents is trained against random agents, to evaluate their capability to learn a game strategy. Second, we deploy a self-play routine that allows each agent to further improve its strategies by playing with evolving versions of itself. Third, once all the agents are trained, we choose the best of them and perform an inter-method competition, where the best agents of each learning method play against each other.

We compare the performance of these agents by measuring the number of wins they have in a series of games, and to better understand and explain their learned strategies, we evaluate their action-selection behavior over time. We explain our results in terms of how the agents learn gaming strategies, and discuss how their specific learning mechanisms affect their learning behavior. 
\begin{figure*}
\begin{center}
  \includegraphics[width=0.75\linewidth]{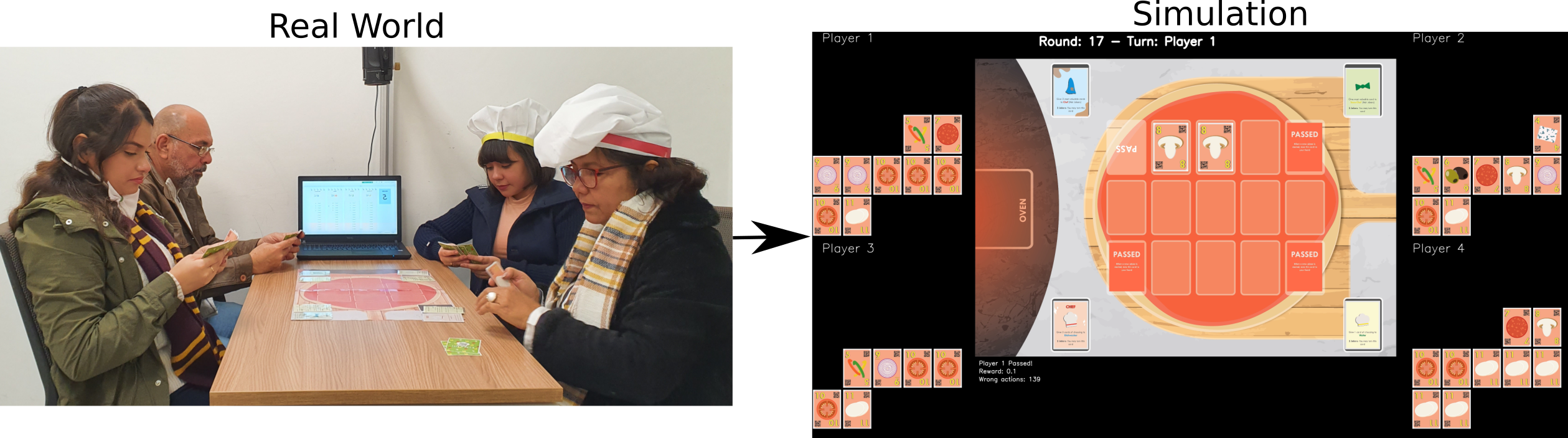}
\end{center}
  \caption{Chef's Hat in real-life gameplay and 1:1 rendered simulation environment.}
\label{fig:gameExample}
\end{figure*}

\section{Learning to be Competitive}
%The concept of competitiveness can be modeled in many different manners. At first, the most important metric for an agent to be competitive is to define the overall environment goal. In our card game scenario, we define the overall metric as winning as many games as possible. This gives each agent a clear goal and allows us to observe how this clear goal affects the agent's behavior while playing the game, and which strategies emerge.

One of the most important metrics for a competitive agent is defining the overall environment goal. In our card game scenario, we define the overall goal as winning as many games as possible. This gives each agent a clear goal and allows us to observe how this affects the agent's behavior while playing the game, and which strategies emerge. We implement the Chef's Hat card game \cite{barros2020food} through the OpenAI-based simulation environment \cite{barros2020chef}, illustrated at Figure \ref{fig:gameExample}. The game represents a controllable action-perception cycle, where each player can only perform a restricted set of actions, and we can directly measure the impact of each action within the game state, and the formation of player's strategy. Furthermore, it allows each player to behave as organically as possible, given the in-game constraints, and allows for a naturally-controllable real-world scenario. 

% In our experiments with artificial agents, we use the Chef's Hat OpenAI-based simulation environment \cite{barros2020chef}, which implements the game in a 1:1 relation. It includes all the complex game mechanics, and allow us to represent all the interaction dynamics.

\subsection{Chef's Hat Card Game Mechanics}

Chef's Hat is played by four players and it has as a theme a kitchen environment. The game was designed, implemented and validated in a way to allow Human-Robot Interaction experiments to be conducted, where one player can be replaced by a robot without changing the game rules or dynamics. %%This contributes to the development of virtual agents playing the game in the same manner as humans, and thus, the simulation allows a true representation of the real-world game scenario.
The game is composed of a role-based hierarchy: each player can either be a Chef, a Sous-Chef, a Waiter, or a Dishwasher. The objective of the players is to be the first one to get rid of their ingredient cards and become the Chef. The player which was most times the Chef is considered the winner of the entire game. The flow of one full game is depicted in Algorithm \ref{alg:ChefsHat}. 

\begin{algorithm}
 Shuffle the deck; \\
Deal an equal amount of cards per player; \\
  Exchange roles; \\
  Exchange cards; \\
   \If{special action is evoked}
  {
    Do special action;
  }
  
 $FirstPlayer\gets Has golden 11$\\
 $FirstPlayer$ discard cards. \\
 
 \While{not end of the game}{
  \For{ each player}{

      \eIf{player can, and want, to discard}{
       discard cards\;
       }{
        pass\;
      }
      \If{All players passed}
      {
        Make the pizza;
         $FirstPlayer\gets Last player to discard$\\
      }
        \If{All players finished}
      {
        End of game.
      }
   }
 }
 \caption{The Game-flow of the Chef's Hat card game.}
 \label{alg:ChefsHat}
\end{algorithm}

During each game there are three phases: Start of the game, Making Pizzas, End of the game. The game starts with the cards having been shuffled and dealt to the players. Then, starting from the second game, the exchange of roles takes place based on the last game's  finishing positions. The player who finished first becomes the Chef, the one that finished second becomes the Sous-Chef, the one that finished third becomes the Waiter and the last one the Dishwasher. Once the roles are swapped, the exchange of the cards starts. The Dishwasher has to give the two cards with the highest values to the Chef, who in return gives back two cards of their liking. The Waiter has to give their lowest valued card to the Sous-Chef, who in return gives one card of their liking. If any player has two jokers at hand, they can perform a special action: in case of the Dishwasher, this is "Food Fight" (the hierarchy is inverted), in case of the other roles it is "Dinner is served" (there will be no card exchange during that game).

Once the cards and roles have been exchanged, the game starts. The goal of each player is to discard all the cards at hand. They can do this by making a pizza, which consists of laying down the cards into the playing field, represented by a pizza dough. The person who possesses the Golden Mozzarella card (with a face value of 11) at hand starts making the first pizza of the game. A pizza is done when no one can, or wants to, lay down any more ingredients. To discard a card, they need to be rarer (i.e. lower face values) than the previously played cards. The ingredients are played from highest to the lowest face value, that means from 11 to 1. Players can play multiple copies of an ingredient at once, but have to always play an equal or greater amount of copies than the previous player did. If a player cannot (or does not want to) play, they pass until the next pizza starts. A joker card is also available and when played together with other cards, it assumes their value. When played alone, the joker has the highest face value (12). Once everyone has passed, they start a new pizza by cleaning the playing field, and the last player to play an ingredient is the first one to start the new pizza.

\subsection{Chef's Hat Card Game Simulation}

To simulate the Chef's Hat game, we implemented our scenario using the OpenAI-based simulation environment of Chef's Hat \cite{barros2020food}. The environment simulates all the game mechanics described above and allows the plugin of different agents to play the game. It comes embedded with dummy agents that randomly perform actions.

The simulator represents the current game state for each player as an aggregation of the cards the player has at hand, and the current cards in the playing field; using a total of 28 values for the state representation, one value per card. For each player, there are a total of 200 allowed actions: to discard one card of face value 1 represents one move, while to discard 3 cards of face value 1 and a joker is another move, and passing is considered another move. Each player can only do one action per game turn.

% Each action taken by a player is validated using a look-up-table, created in real-time based on the player's hand and the cards in the playing field. This is a crucial step to guarantee that a taken action is allowed given the game context and to guarantee that the game rules are maintained.  Figure \ref{fig:possibleActions} illustrates an example of calculated possible actions given a game state. The blue areas mark all the possible action states, while the gray areas mark actions that are not allowed due to the game's mechanics. 

\subsection{Learning to be the Chef}

To train artificial agents to play Chef's Hat, we employ different reinforcement learning algorithms based on Q-learning. Q-Learning allows our agents to apply a temporal difference calculation when updating the policy network to maximize the state transitions that will lead to the optimal reward. In this regard, Q-learning showed a faster convergence and a simplified learning process \cite{shoham2003multi, gosavi2009reinforcement, kiumarsi2017optimal} when compared to other reinforcement learning methods.

\textbf{Chef's Hat Greedy Policy.} To provide an ideal balance between exploration and exploitation, an $\epsilon$-greedy exploration mechanism is usually adopted \cite{tokic2010adaptive, painter2012greedy, efroni2018multiple}. In the traditional form, each agent implements an action selection mechanism that ensures an exploration through random action selection at the beginning of the training:

\begin{equation}
a_t = \left\{\begin{matrix}
random(a) & if & x \leq \epsilon\\ 
 Network(state))& if & x > \epsilon  
\end{matrix}\right.
\end{equation}

where \emph{random(a)} represents a random action selection over the entire action space,  $\epsilon$ is the greedy factor, and $x$ a random number selected at each action. Usually $\epsilon$ starts with a higher value at the beginning of the training and it is reduced each time the policy is updated. This guarantees that the model performs a large number of exploratory steps at the beginning of the learning but incrementally starts to trust more and more on the policy update by the end of the learning phase.

In our scenario, however, performing a fully random action is not beneficial to the agent. As the game simulation only allows for valid actions to go through, and thus moving on towards the next game state, choosing random actions could lead to an agent getting stuck in a state until it chooses randomly a valid action. At the same time, penalizing the agent for choosing an invalid action is not ideal as it creates an unbalanced training set with a reward representing multiple goals: win the game, and perform valid actions, which creates unstable and unfocused learning. To solve this problem, we introduce here an updated greedy policy for the Chef's Hat agents. Instead of selecting a random action, the agent calculates the allowed actions given a state using Algorithm \ref{alg:possibleAction}.

\begin{algorithm}
 
 $PossibleActions\gets []$\\
 $JokersOnBoard\gets calculateJokersOnBoard()$\\
 $CardsInHand\gets cardsInHand()$\\
 $CardsInPlayingField\gets cardsInPlayField()$\\
  
 $firstAction \gets isFirstActionOfTheGame()$ \\
  
   \For{$CardValue\gets11$}{
        \For{$CardQuantity\gets11$}{
             \eIf{$CardValue$ in $CardsInHand$ and $CardValue \geq Max(cardsInPlayField)$}{
             
                 \eIf{$CardQuantity \leq CardsInHand[CardValue]$ and in $CardQuantity \geq Max(currentBoard) + JokersOnBoard$}{      
                 
                     \eIf{is $firstAction$}{
                         \eIf{$CardValue == 11$}{
                             $PossibleActions \gets append(1)$
                         }
                         {
                           $PossibleActions \gets append(0)$
                         }
                     
                      }
                      {
                         $PossibleActions \gets append(1)$
                      }
    
                }
                {
                     $PossibleActions \gets append(0)$
                }

            }
            {
                 $PossibleActions \gets append(0)$
            }

         }
   }
    \KwResult{$PossibleActions$}
 \smallskip
 \caption{Chef's Hat novel greedy action selection algorithm. It creates a vector containing all the 200 possible actions and which of them are allowed given a certain state.}
 \label{alg:possibleAction}
\end{algorithm}

The output of Algorithm \ref{alg:possibleAction} is hot-encoding with all the 200 possible actions, with a 1 representing an allowed action and a 0 representing an invalid action given that specific state. Our $\epsilon$-greedy function is then represented by:

\begin{equation}
a_t = \left\{\begin{matrix}
random(PossibleActions(state)) & if & x \leq \epsilon\\ 
 Network(state) & if &x > \epsilon  
\end{matrix}\right.
\end{equation}

\begin{figure}
    \centering
    \includegraphics[width=0.7\linewidth]{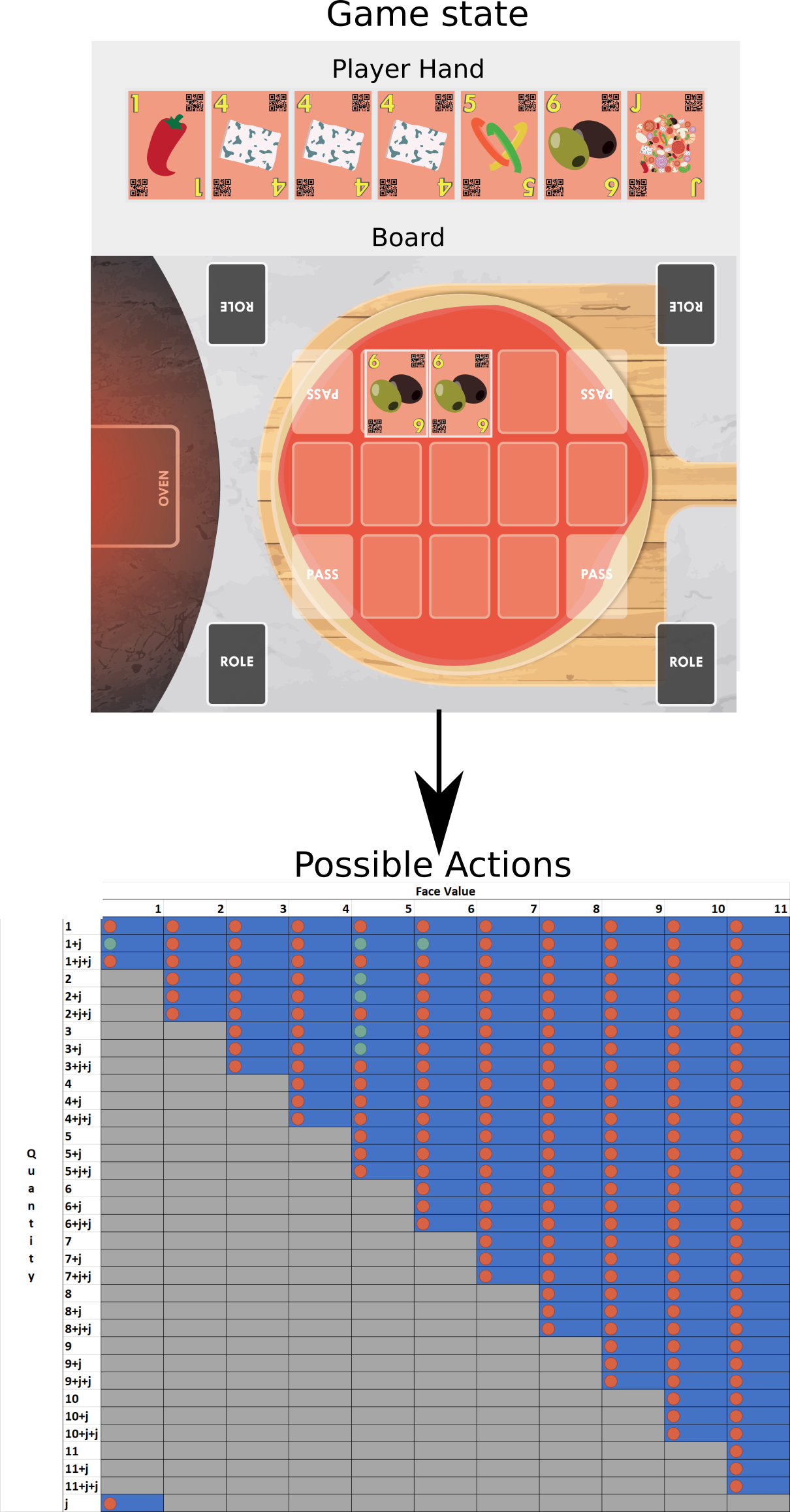}
    \caption{Example of possible actions given a certain game state. The columns represent the card face values, and the rows represent the number of cards to be discarded. The letter "j" represents the presence of a joker. It marks all the allowed actions within the game mechanics (blue regions) and not allowed actions (the gray regions), and the currently allowed actions (green dots) given a certain state.}
    \label{fig:possibleActions}
\end{figure}

To better understand the output of Algorithm \ref{alg:possibleAction}, Figure \ref{fig:possibleActions} illustrates an example of calculated possible actions given a game state. The blue areas mark all the possible action states, while the gray areas mark actions that are not allowed due to the game's mechanics. The green dots illustrate the possible actions given that specific state, and the red dots display the invalid actions.

\subsection{The Tale of Three Learners}

In order to validate our learning scenario and the Chef's Hat greedy action selection mechanism, we adapted three popular Q-learning-based methods: Deep Q-Learning - DQL \cite{van2016deep}, Advantage Actor-Critic - A2C \cite{mnih2016asynchronous}, and Proximal Policy Optimization - PPO \cite{schulman2017proximal}. Each of these algorithms represents one particular aspect of reinforcement learning, and our goal is to demonstrate how they learn and behave when deployed in our scenario using our specific greedy action selection process.

 For each action taken by an agent, we calculate a mask composed of the output of Algorithm \ref{alg:possibleAction}. This mask is applied to the output layer of the neural network that calculates the Q-values of the actions for each algorithm. The mask is extremely important to guarantee that the outputs of the networks are in agreement with the games' mechanics, and thus, focus the Q-values maximization towards finding the best game-play strategy.

All learning agents parameters, illustrated in Figure \ref{fig:models}, were optimized using a TPE optimization implemented by the Hyperopt \cite{bergstra2013hyperopt} library. Each of the learning agents implemented a single optimization routine for minimizing its loss when playing against dummy random agents. We implemented the agent using the Keras library \cite{gulli2017deep}, and our agents and experiments implementations are publicly available \footnote{https://github.com/pablovin/ChefsHatGYM}.

\subsubsection{Deep Q-Learning}

% \begin{figure}
%     \centering
%     \includegraphics[width=0.4\columnwidth]{DQLNetwork.png}
%     \caption{The DQL-based agent with a single network that learns to associate Q-values to a certain state.}
%     \label{fig:dqlNetwork}
% \end{figure}

Deep Q-learning is an evolution of the standard Q-learning method and introduces two novel aspects: a target model and the experience replay. The target model helps to stabilize the learning of Q by providing a stable Q-estimation over the training. The experience replay stores the agent's own experience by saving important steps taken by the agent, to increase the available data for learning state/action pairs through batch-learning. Deep Q-learning has been recently applied to teach agents to play complex video games with great success \cite{hausknecht2015deep, hester2018deep, meng2019qualitative}, mostly due to their capability of performing batch-learning using the experience replay. This increases drastically their training time but results in finding optimal game-winning strategies. We expect to see this behavior reflected on how this agent learns different strategies to play our game as well. 
% Figure \ref{fig:dqlNetwork} illustrates the final architecture of the DQL agent.

% The target model introduces a second, the target model, policy which receives a snapshot of the original policy after a certain number of training steps.

% The target model is used to obtain the target Q learning when calculating the $TD_dql$:

% \begin{equation}
% TD_dql =r_t \times \gamma \times maxQ\left (s__{t+1, a_t}  \right ) - maxQ_t\left (s__{t1, a_t}  \right )
% \end{equation}

% \noident where $maxQ_t\left (s__{t1, a_t}  \right )$ is the Q-values obtained from the target network.

% The experience replay stores the agent's own experience by saving important steps taken by the agent, to increase the available data for learning state/action pairs through batch-learning. Deep Q-learning was recently applied to teach agents to play complex video games with great success \cite{hausknecht2015deep, hester2018deep, meng2019qualitative}. In our scenario, Q-learning has an advantage of updating the agent towards winning the game, and not only towards taking single-stepped actions. Figure \ref{fig:dqlNetwork} illustrates the final architecture of the DQL agent.

\subsubsection{Advantage Actor-Critic}

% \begin{figure}
%     \centering
%     \includegraphics[width=0.4\columnwidth]{A2CNetwork.png}
%     \caption{The A2C-based network presents a two-tailed implementation that encodes the state with a shared representation and decodes the Q-values and critic values using two separate outputs.}
%     \label{fig:a2cNetwork}
% \end{figure}

\begin{figure}
    \centering
    \includegraphics[width=1\columnwidth]{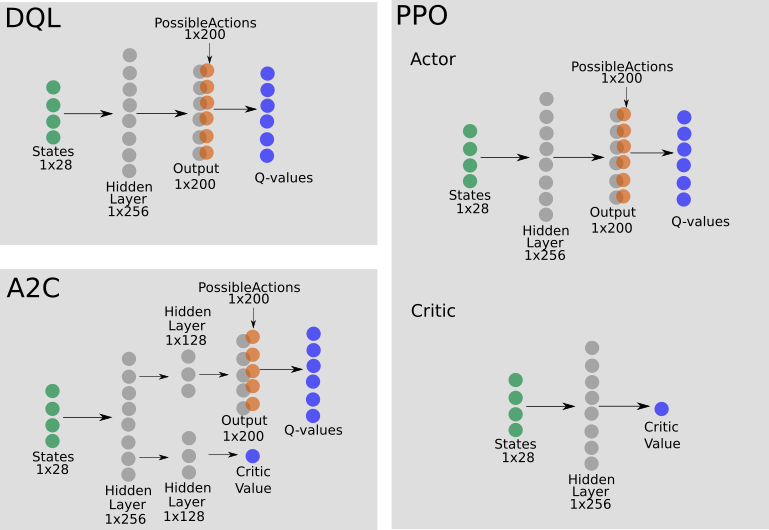}
    \caption{The detailed implementation of our three agents: DQL, A2C, with a double-tail implementation, and PPO with individual actor and critic networks.}
    \label{fig:models}
\end{figure}

Actor-critic models present a hybrid learning method where an agent learns how to estimate the Q-values for a given state by following policy, the actor-network, and updates the chosen Q-values importance by a value-function approximator, the critic network. Advantage Action-critic \cite{mnih2016asynchronous} was introduced recently to stabilize the learning of the two networks by introducing the advantage function, which helps the entire model to identify, given a certain state, how much better it is to take a specific action compared to an average of all the actions. Recent research demonstrates how A2C models present stable learning for video-games scenarios \cite{clary2019let}, and we expect to observe a steady improvement of this agent while learning a strategy. Our implementation of the A2C model uses a common decoder and a two-tailed network architecture and it is represented in Figure \ref{fig:models}. 

% Typical Action-Critic models are a on-policy learning methodology that learns how to measure a value for a policy while following it. This allows a stable online learning mechanism, where the model outputs the Q-values for an action, actor network, and independently, estimates a critic value about the actions selected by the actor using a critic network.

% \begin{equation}
% A(s_t,a_t) = r_t + \gamma \times V(s') - V(s)
% \end{equation}

% \noident where $V(s)$ represents the critic value for a given state, and $V(s')$ for the future state state. The advantage function is used to estabilizes training the actor network, while the critic network uses the discounted rewards as target. 

\subsubsection{PPO}

% \begin{figure}
%     \centering
%     \includegraphics[width=0.4\columnwidth]{PPONetwork.png}
%     \caption{Our PPO-based agent implements independent actor and critic networks, both receiving the state as input.}
%     \label{fig:ppoNetwork}
% \end{figure}

Our third implemented learning model is the Proximal Policy Optimization (PPO) \cite{schulman2017proximal}. PPO is a recently introduced policy-based method, which follows the same learning structures as the A2C. It, however, implements an adaptive penalty control, based on the Kullback–Leibler divergence, to drive the updates of the agent at each interaction. This allows the model to create an update region that functions similarly to the stochastic gradient descent optimization, simplifying the necessity of the algorithm to keep large memory-replays or complex update rules. PPO has been used recently with great success in different competitive scenarios, where the environment is constantly changing  \cite{bansal2017emergent, kidzinski2018learning}. We expect that our PPO agent will present quick adaptation to newly perceived competitive strategies, in particular when playing against the other agents, which have a slower adaptation mechanism. The figure illustrates our PPO agent. 
% Figure \ref{fig:ppoNetwork} illustrates the final model of our PPO agent.

\section{Evaluating Competition}

The goal of this paper is to demonstrate, evaluate and understand how the three reinforcement learning methods described above behave when learning in a multiplayer competitive scenario provided by the Chef's Hat simulation environment. As such, we separate our evaluation routines into three experiments: First, we train one agent implementing each of these methods playing against three other agents implementing random action selections. Second, we perform a self-play training routine where each of the learning agents plays with different generations of themselves. Finally, we choose the best learning agent from the self-play experiments and play a competitive game with the three agents and a random agent.

\textbf{Reward and Metrics.} To train our agents we use an overall rewarding strategy: The environment gives a full reward (1.0) when performing the action that leads an agent to win the game. Every other reward is set to -0.01 to promote exploration within the agent Q-learning algorithm in order to avoid an unoptimal solution. Given the temporal-difference learning, the agent will learn how to generate strategies, composed of a sequence of actions, in order to achieve the maximum reward without receiving any prior information from the environment. 

For each experiment, we evaluate the agent's performance by calculating the average of victories for all the games the agent played in a series of 10 experimental runs of 100 games each, totaling 1000 games. To help us understand and explain how the agents learn, we also calculate the selected action Q-values over all the games, which will give us an insight on how is the agents confidence in selecting certain actions during the game-play. We post-process all the Q-values of an agent, per turn, using a softmax function, which help us to exhibit the Q-values as a probability, improving readability.

To fully illustrate our experimental setup, we report all the experiments, training and validation routines, agent combinations and the number of games in Table \ref{tab:experimentalSetup}.
 
%  Finnaly, to illustrate we calculate the summed reward over the entire experiment, which give us an overview of the overall performance of the agent over the entire game-span, combining victories and number of rounds. The total number of victories will indicate how well the agent is able to deal with the strategies of the other players. 
\begin{table}[h!]
\centering
\begin{tabular}{ c | c | c | c }

% \noalign{\smallskip}\hline
% \multicolumn{5}{c}{\cellcolor[gray]{0.9}\emph{vs. Random}} \\
% \hline\noalign{\smallskip}

 \cellcolor[gray]{0.9}\emph{Exp.} &  \cellcolor[gray]{0.9}\emph{Routine} & \cellcolor[gray]{0.9}\emph{Agents} &  \cellcolor[gray]{0.9}\emph{\# Games}\\ \hline
 \multirow{ 6}{*}{Random}  & \multirow{ 3}{*}{Train} & $1\times$ DQL vs $3\times$ Random & 1000 \\
   &&$1\times$ A2C vs $3\times$  Random & 1000 \\
   &&$1\times$ PPO vs $3\times$  Random & 1000 \\ \cline{2-4}
      & \multirow{ 3}{*}{Val.} & $1\times$ DQL vs $3\times$  Random & $10\times100$ \\
   &&$1\times$ A2C vs $3\times$  Random & $10\times100$ \\
   &&$1\times$ PPO vs $3\times$  Random & $10\times100$ \\ \hline
   
  \multirow{ 6}{*}{Myself}  & \multirow{ 3}{*}{Train} & $4\times$ DQL & $50\times1000$ \\
   &&$4\times$ A2C & $50\times1000$ \\
   &&$4\times$ PPO & $50\times1000$ \\ \cline{2-4}
      & \multirow{ 3}{*}{Val.} &  $DQL_1$ vs $DQL_25$ vs $DQL_{50}$ vs $DQL_{r}$ & $10\times100$ \\
   &&$A2C_1$ vs $A2C_25$ vs $A2C_{50}$ vs $A2C_{r}$ & $10\times100$ \\
   &&$PPO_1$ vs $PPO_25$ vs $PPO_{50}$ vs $PPO_{r}$ & $10\times100$ \\ \hline 
   
     \multirow{ 2}{*}{Others}  & Train. & DQL vs A2C vs PPO vs Random & 1000 \\\cline{2-4}
      & Val. &  DQL vs A2C vs PPO vs Random & $10\times100$ \\ \hline 
  
%  Dense layer & 256 & ReLU \\
%  Dense layer & 200 & Tanh (output layer) \\\hline
%  Training Parameter & \multicolumn{2}{c}{Value}  \\ \hline
%  Batch Size & \multicolumn{2}{c}{ 32} \\
%  Memory size & \multicolumn{2}{c}{ 2000} \\  
%  discount rate & \multicolumn{2}{c}{ 0.95}  \\ 
%  exploration decay & \multicolumn{2}{c}{ 0.995} \\\hline   
 
\end{tabular}
\caption{Experimental setup: training and validation routines, agent combinations, and number of performed games per routine.}
\label{tab:experimentalSetup}
\end{table}
\subsubsection{vs Random}

Our first experiment puts each of the learning agents to play against three dummy agents. We perform a training routine that lasts 1000 games. We then perform an evaluation routine where each trained agent plays 10x100 games against the random agents, without further training, and we measure the average of total of victories achieved by each agent per 100 games together with the standard deviation. This experiment aims to give us important information about how each trained agent learns to beat a simple strategy based on random selections. 

% Separating the training routine with a testing routine allows us to validate the learned strategies by the agents.

% The random agents are implemented to provide a random action selection based on the possible actions for each given state. Each learning agent is trained from scratch for 3000 games, meaning that at the beginning of the learning routine they will provide mostly random actions. Once they start learning, their behavior change. After each experiment, we run a test scenario where the trained agent plays 1000 games against the random agents, but without training. This allows us to validate the effectiveness of the learning procedure. The goal of this experiment is to demonstrate how each of the agents learns to devise winning strategies against random agents. 

\subsubsection{vs Myself}

Our second experiment is composed of a self-playing routine. For each self-play generation, we train agents playing against each other for 1000 games. In order to increase the oponents variability and avoid an overspecification of the agent, in every generation, we save the best and second-best agents in a list, based on their averaged summed reward when playing against each other in a validation routine composed of 1000 games without further training. For the next generation, we copy the best agent from the previous generation and put it to play against three other agents, which can be pulled from the best and second-best list, a newly instantiated agent, or a random agent. The selection happens randomly, with the same probability of choosing any of these agents. We repeat the self-play routine for 50 generations, totaling 50.000 played games per learning method. We evaluate the impact of the self-playing routines by getting the first, the 25th and the last generation to play a game against the best agent from the previous experiment for 10x100 games, and measure the averaged number of victories and standard deviation. This experiment allows us to observe how the self-playing routine affects the trained agents' performance within different generations.
% For each self-play generation, we train agents, with the same learning method, playing against each other for 1000 games. We then calculate the best agent by measuring their summed reward and re-start the routine by replacing the agents with the best agent. 
% We repeat this procedure for 10 generations, always copying the best agent of the previous generations to the new one.
% This experiment is composed of a self-play routine: we start the game with four agents implementing the same learning algorithm. After an initial experiment with 1000 games, we select the best agent, based on its summed reward output and start a new generation of agents which are initialized as a copy of the best agent. We run this experiment for 10 generations and provide a broad comparison of the performance of the agents' in-between generations.

\subsubsection{vs Others}

Our last experimental setup involves an inter-method evaluation. We take the best-trained agents for each learning method, based on the results of the \emph{vs. Myself} experiments, and put them to play against each other and a dummy agent. To play agains the dummy agent will normalize their behavior by providing a super easy agent that all of them can beat. We perform two evaluation routines here. The first involves these agents playing against each other for 10x100 games, without further training. The second instead consists of a training routine that lasts 1000 games followed by an evaluation routine that lasts 10x100 games without training. We calculate here the average victories for each agent, together with the standard deviation. This experiment will exhibit the performance of the implemented agents when compared to each other, and how they can adapt to more complex strategies than random action selection.

%\subsection{Experimental Summary}

\section{Results}
The results from all three experiments - \emph{vs. Random}, \emph{vs. Myself} and \emph{vs. Others} - are depicted in Table \ref{tab:results}.

\begin{table}[h!]
\center
\begin{tabular}{ c c c c c } 

\noalign{\smallskip}\hline
\multicolumn{5}{c}{\cellcolor[gray]{0.9}\emph{vs. Random}} \\
\hline\noalign{\smallskip}

Model & Victories & Random1 & Random2 & Random3 \\ 
\noalign{\smallskip}\hline\noalign{\smallskip}
DQL & \textbf{66.8} $\pm$5.69	& 9.7 $\pm$3.13	& 12.9	$\pm$4.66 & 10.6	$\pm$1.8\\ 
A2C & \textbf{65.1} $\pm$5.19	& 9.3 $\pm$3.1 & 12.1	$\pm$4.35 &	13.5	$\pm$3.58\\ 
PPO &	\textbf{83.1}	$\pm$4.18	& 4.7 $\pm$2.19 & 6.0 $\pm$2.28	& 6.2 $\pm$1.83 \\ 
						
\noalign{\smallskip}\hline
\multicolumn{5}{c}{\cellcolor[gray]{0.9}\emph{vs. Myself}} \\
\hline\noalign{\smallskip}

Model & Gen-1 & Gen-25 & Gen-50 & Random \\ 
\noalign{\smallskip}\hline\noalign{\smallskip}
DQL & 19.4	$\pm$4.78 & 24.8	$\pm$4.98& \textbf{42.9}	$\pm$7.06 & 12.9	$\pm$6.64\\
A2C & 25.4	$\pm$4.39 & 29.1	$\pm$6.14 & \textbf{34.5}	$\pm$7.12 & 11	$\pm$2.86\\
PPO &16.9	$\pm$3.36 & 32.5	$\pm$3.75 & \textbf{40.3} $\pm$3.52 & 10.3	$\pm$4.1\\

\noalign{\smallskip}\hline
\multicolumn{5}{c}{\cellcolor[gray]{0.9}\emph{vs. Others}}\\
\hline\noalign{\smallskip}

Model & \multicolumn{2}{c}{Before training} & \multicolumn{2}{c}{After training}\\
\hline\noalign{\smallskip}

DQL & \multicolumn{2}{c}{35.9	$\pm$3.11}  &	\multicolumn{2}{c}{35.9	$\pm$3.11} \\
A2C & \multicolumn{2}{c}{18.9	$\pm$3.51} &	\multicolumn{2}{c}{4.9	$\pm$2.84} \\
PPO & \multicolumn{2}{c}{\textbf{42.8}	$\pm$5.06} & \multicolumn{2}{c}{\textbf{48.5} $\pm$40.6} \\
Random & \multicolumn{2}{c}{2.4	$\pm$0.8} &	\multicolumn{2}{c}{3.3	$\pm$1.85} \\

\noalign{\smallskip}\hline\noalign{\smallskip}
\noalign{\smallskip}
\noalign{\smallskip}
\end{tabular}
\caption{Results for all three experiments.}
\label{tab:results}
\end{table}

% \begin{figure}
%     \centering
%     \includegraphics[width=1.0\columnwidth]{exp1Results.png}
%     \caption{Averaged victories per 10 series of 100 games when validating each of the agents after training them against random opponents.}
%     \label{fig:exp1Results}
% \end{figure}

\subsection{vs Random}
We observe that the PPO agent achieves highest number of victories during the validation routine with an average of 83.1 victories per 100 games, followed by DQL (66.8 averaged victories) and A2C (65.1 averaged victories) respectively. As these experiments were performed while playing against random agents, these numbers inform us that all the agents learned how to beat a random strategy, with the PPO agent been the best on it.
% The Q-values evolution during training gives us an insight on when, during training, these agents start to select actions with higher confidence. While A2C and PPO agents take some games to provide higher confidence in the selected action, the DQL agent achieves a faster increase in confidence, starting already from the first games. This is probably due to the experience of replay-based training, which increases the number of state/action pairs that this agent is trained on per training routine. 

\subsection{vs Myself}

Our results from the self-playing experiments clearly show that the \emph{vs Myself} agents learned how to beat the strategies learned by the \emph{vs Random} agents. What is important to notice is the higher standard deviation obtained by the DQL and the A2C agents when compared to the PPO agent. Again, given the PPO advantage on fast adapting, it presents a much more consistent behavior on learning the best strategies to play against a more varied type of opponents. Also, our results validate our training routine by having the final generation of all the agents always achieving more victories than the previous ones. 

% \begin{figure}
%     \centering
%     \includegraphics[width=1.0\columnwidth]{exp2Results.png}
%     \caption{Averaged victories per 10 series of 100 games when validating each of the agents after the self-playing training routine.}
%     \label{fig:exp2Results}
% \end{figure}

\subsection{vs Others}

Our third and last experiment put the best agent from each learning algorithm (based on the results of the \emph{vs Myself} experiment) to play against each other. This result, illustrated in the "Before Training" column, shows us that the PPO agent is the one with the best performance, followed closely by the DQL agent, and both with much better results than the A2C. The A2C number of victories tell us that the strategies it learned were much less successful when compared to the PPO and DQL. 

This is much clearer when we re-train the three agents, making them adapt to each other strategy (illustrated in the "After Training" column). The re-adaptation causes the DQL and the PPO agent to obtain similar performance, with a slight advantage to PPO, while the A2C agent seems to be completely ineffective against the other two. This can be explained by how these agents learn. The fast adaptation from the PPO agent presents an expected advantage compared to the A2C agent, while the experience replay from the DQL helps it to experience many more training samples, and to focus on learning a set of winning strategies. This behavior is better explored and explained in the next section.

% \begin{figure}
%     \centering
%     \includegraphics[width=0.8\columnwidth]{exp3Results.png}
%     \caption{Averaged victories per 10 series of 100 games when validating each of the agents before and after training them against themselves.}
%     \label{fig:exp3Results}
% \end{figure}

\section{What is to be competitive?}

Calculating the overall number of victories per agent tells us if they were successful in maximizing the goal of the game. However, once we prove that these agents can learn, and some better than the others, it is of high importance to shed a light on how they achieve such performance in the competitive Chef's Hat scenario. In this regard, we discuss below our interpretations of how these agents learn the game strategy and how they learn to be competitive when playing against each other.

% Each agent learns to play the game by receiving a feedback based on the reward they obtain from the environment. The feedback is used to update the agent's capability of choosing an action that will lead to winning the game. For each state, in our case a game turn, the agent calculates the Q-values for each of the 200 allowed actions, and chooses the highest one, which represents the one with the higher probability of achieving a victory, given the current game state.

\subsection{How do I learn an action-selection strategy?}

\begin{figure}
    \centering
    \includegraphics[width=1\linewidth]{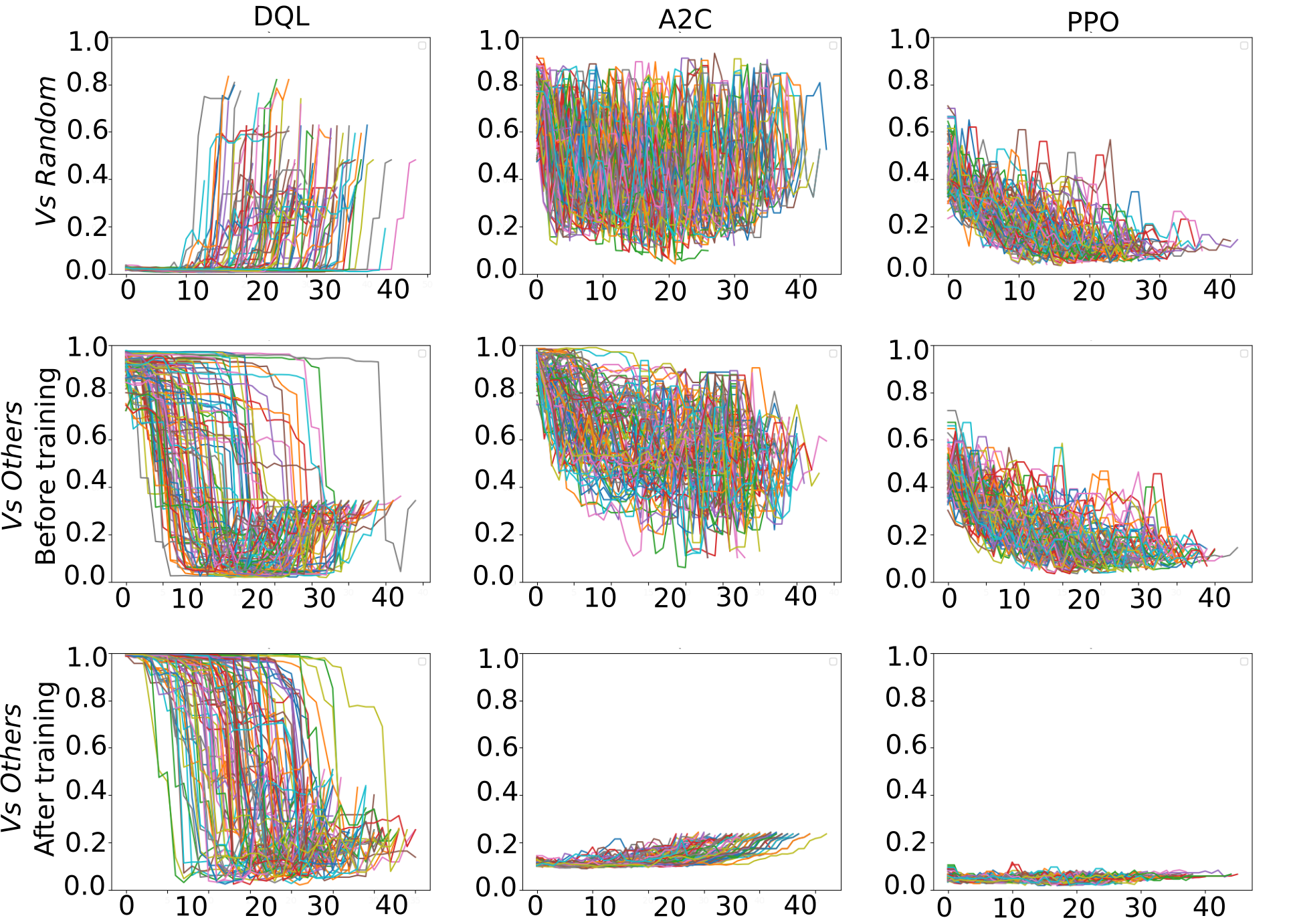}
    \caption{Q-values readings (Y axis) for each action of a game (X axis) for a hundred games following the \emph{vs. Random}, \emph{vs. Others} before and after training routine.}
    \label{fig:discussion2TrainingQValues}
\end{figure}

To have a better insight on the learned action-selection strategy per agent, we run a hundred games in the \emph{vs. Random} and vs \emph{vs. Others} before training and \emph{vs. Others} after the training routines and plot the selected Q-Values over all the played matches in Figure \ref{fig:discussion2TrainingQValues}. In the \emph{vs. Random} games, we keep the random agents receiving the same card distribution when playing against the random agents, to reproduce a similar initial condition. 

We observe that the \emph{vs. Random} routine, the DQL have higher confidence on a single selected-action usually by the end of the game. This possibly indicates that this agent learned a small set of actions that guarantee it a win against the random agent usually be the mid-end of a game-play. The A2C agent has a more distinguished action-selection pattern, where a single action seems to have high-confidence over the entire game-play. The PPO behaves somehow the opposite of the DQL agent, as it presents higher confidence in a single action at the beginning of the match while having devised different strategies, demonstrated by having low confidence in a single action, through the duration of a game.

% the A2C agent presents focuses on a few actions always. This behavior allows the A2C agent to beat a random agent, identifying that this agent learns a single strategy to beat them. All the agents, however, present good performance when playing against the simple random agents, which validates their own strategy to beat them.

When playing a game against each other, on the \emph{vs. Others} before training routine, the A2C and PPO somehow present the same behavior as the vs Random training. As this scenario is composed of the agents that learned via self-play, we can infer that their action-selection strategy was not much altered by this training routine. Also, this scenario differs from the \emph{vs. Random} scenario by providing a much more complex and dynamic state throughout the game as each action taken by an agent has a direct impact on its opponents. This is reflected directly in the action-selection behavior of the agents changing through time. The DQL agent, however, changed its behavior drastically. It seems to have higher confidence in a single action at the beginning of the game, an opposite behavior from the \emph{vs. Random} routine. This behavior change can be explained by the batch-learning technique used by the DQL. Probably on the \emph{vs. Random} scenario, the agent learned a set of similar strategies to beat the random agents and reinforced it. On the \emph{vs. Others} routine, the agent learned another set of few strategies that seem to win most of the games, which is reflected in the agent's behavior change.

% This indicates that the DQL agent, after training for 50 thousand games, was able to change completely its 

% The strategy depicted by the A2C agent's Q-values reading tells us that it still focuses on trusting in a very few numbers of actions, given the current game state. Even though the other agents are changing their own behavior through the gameplay, the A2C's behavior does not change. S

% imilar behavior happens on the first part of the match for the DQL agent, while by the end of the match it shows an adaptation pattern where the confidence on different actions emerges. The PPO agent seems to not have a specific predilection to certain actions and somehow is able to consider different actions for each game state, in particular by the end of the match.

After training the agents on the \emph{vs. Others} after training routine, the A2C and PPO agents change their behavior. In this scenario, the agents were updated to win the game by playing against each other, and thus, the update routine rewarded behavior that hinders the other players to win. That means they probably try to learn strategies to counter each other's playing style. The A2C agent predilection to a specific action seems to disappear, and it presents a behavior similar to the PPO agent in the previous experiments. However, this does not translate onto victories, on the opposite, based on our results, it seems that the A2C agent becomes much more ineffective. This probably indicates that the A2C agent probably is lost, and did not learn any strategy to play against the other two. The PPO agent seems to continue its adaptation towards different actions per game-state, which translates in the highest number of victories.
The DQL agent keeps the same behavior as the previous experiment, which seems to help it to win games. It shows a slight delay on when to focus on a single action. Probably it learned a single strategy that seems to be quite effective against the other players. This probably indicates the difference between the DQL and PPO agent, while the DQL agent learn a set of few strategies that win the game, the PPO agent learned a more balanced game-play style and different strategies to counterbalance the other agents behavior.

% Q-values peak by the end of the game, with a couple actions (represented by different colors on the plot) having the highest value. This indicates that the DQL agent cannot identify a specific action-selection strategy before the end of the game. This differs from the PPO agent, which has a Q-value peak in the beginning of the game, indicating that it probably plays using a action-selection strategy to set the game pace from the beginning. The A2C agent shows different peaks of confidence throughout the entire game, with many different actions having a similar confidence value. This is probably due to its not learning specific action-selection strategies to win the game, but somehow trying to cope with the other players dynamics. It showed to be enough to win against the random agents, but not enough to win against the DQL and PPO agent.

% While this behavior happens on all thee evaluation routines, it is clear that on the vs Everyone before training routine, the confidence values are very small when compared to the other two routines. When training against each other, the confidence increases, indicating that the agents did adapt to each other's way of playing.

\subsection{How do I learn to be competitive?}

% Each agent learns to play the game by receiving a feedback based on the reward they obtain from the environment. The feedback is used to update the agent's capability of choosing an action that will lead to winning the game. For each state, in our case a game turn, the agent calculates the Q-values for each of the 200 allowed actions, and chooses the highest one, which represents the one with the higher probability of achieving a victory, given the current game state.

\begin{figure}
    \centering
    \includegraphics[width=1\columnwidth]{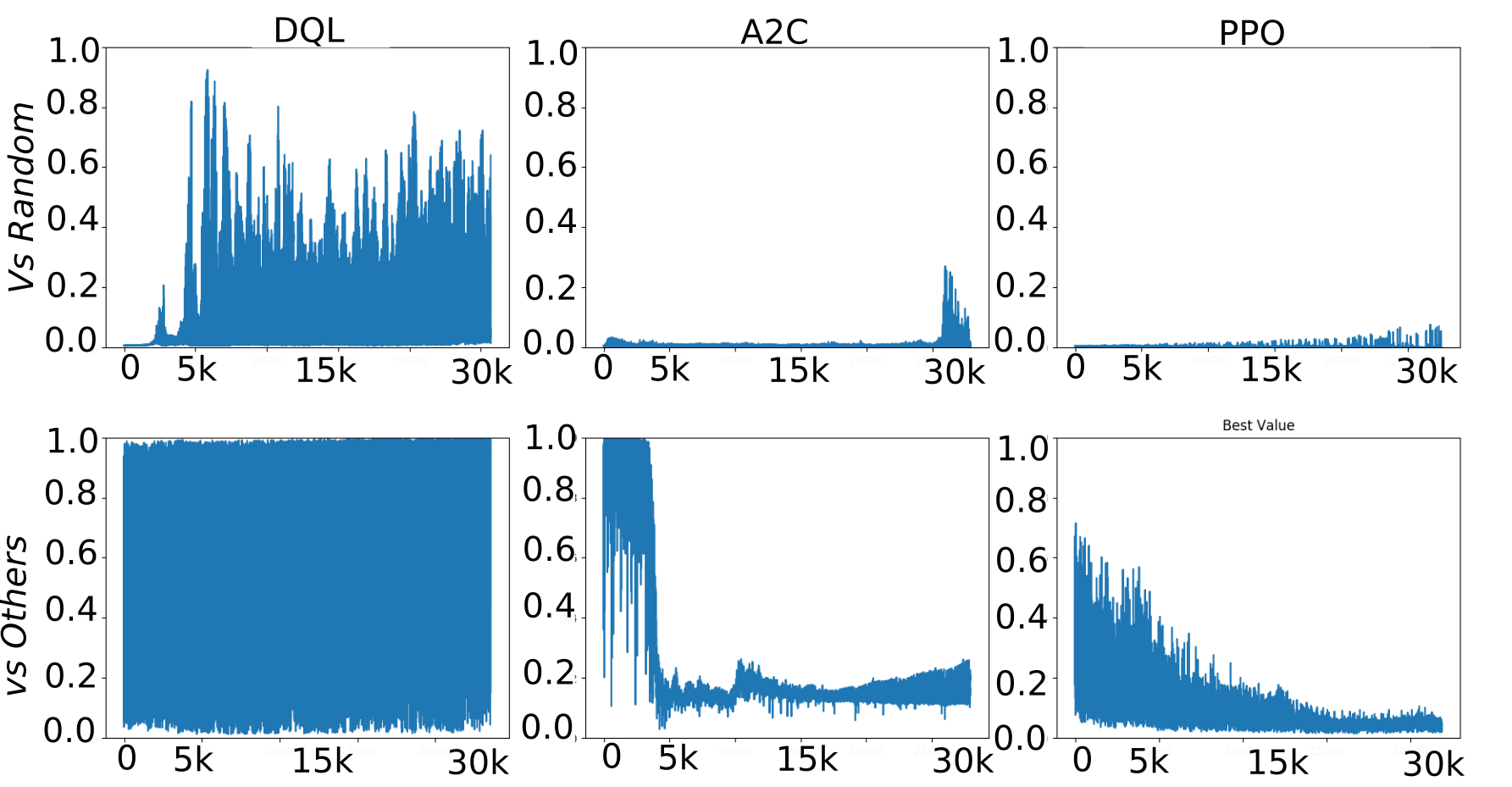}
    \caption{Q-values evolution (Y axis) for each action (X axis) taken in a 1000 games training routine following the \emph{vs. Random}, and \emph{vs. Others} routine.}

    \label{fig:discussion1TrainingQValues}
\end{figure}

Observing the evolution of the selected Q-value during training against the random agents, illustrated in Figure \ref{fig:discussion1TrainingQValues}, gives us an insight into how the learning algorithms devise strategies to beat their opponents. 

When training against the random agents, the DQL agent presents a quick growth on Q-values at the beginning of the training routine, which can indicate that it associates the selected action with a higher chance of achieving the maximal goal, i.e. winning the game, very early on. This corroborates with our insights on the single-game observation, and it is an indication that the agent learns fairly early during training a specific set of strategies to be followed with high confidence.

This behavior changes in the A2C and PPO agents. They take longer to show an increase in the selected actions Q-values, which indicates they need more time to establish a state-action association with confidence while playing the game. They present lower confidence in selecting a specific action when compared to the DQL agent, which however does not translate onto less overall victories. In the case of the PPO, this is exactly the opposite behavior. This can be explained as the PPO agent learning steadily how to play against a random player, and in this way deriving many different strategies, instead of a strict set as the DQL. In this regard, it provides the best performance probably due to its adaptive learning mechanism, when compared to A2C.

When training in the \emph{vs. Others} experiment, the behavior changes. The DQL agent maintains high confidence during the entire training. The A2C agent appears to decrease its confidence drastically mid-training routine, which corroborates with our understanding that it loses focus and it is not able to devise winning strategies. The PPO agent shows an interesting behavior, as it reduces its own confidence in one single action over the training routine. This is probably due to its fast adaptation on finding different strategies to beat the opponents over time, and learning a high number of associations between game-states and actions. 

All the agents' behavior reflects directly their learning mechanisms. The memory replay of the DQL makes it focus on a specific action, probably reinforced by the actions inside the memory themselves. We believe that with more games, the DQL agent would probably learn different strategies as its memory would grow over time. The PPO fast adaptation translates into associating more connections between action and states than the other two algorithms, in particular when training against each other. The A2C struggles to keep the pace of the PPO, and without the focused reinforcement that the DQL has, it loses its ability to adapt quickly, translating to the smallest number of victories.

\section{Conclusion}

In this paper we presented a broad experiment with three different reinforcement learning algorithms playing the competitive Chef's Hat card game. We implemented these algorithms in agents and trained them to play the game. To evaluate the agents, we performed three validation routines - playing against random agents, self-playing, and playing against each other. We described how each learning algorithm behaved within the competitive scenario, and how their learning characteristics contributed to their performance. The PPO-based agent presented the best performance in all of our tasks, demonstrating how its quick update mechanisms contributed to competitive learning.

The agents learned different action-selection strategies, and their learning nature affected the way they tried to optimize their gameplay style. From our results, we consolidated the Chef's Hat card game simulation environment as a challenging task to be learned, and set the initial work on understanding how reinforcement learning can be used in such a competitive task.

We envision the development of further specific adaptations to reinforcement learning agents to be more competitive in the Chef's Hat card game. Given the proximity to the real-world card game scenario, we also encourage further research on applying such agents to play against real humans and embodied agents, such as social robots.


%\section*{Acknowledgment}




%\bibliographystyle{IEEEtran}
\balance
\bibliographystyle{IEEEtran}
\bibliography{bib}

\end{document}